%% file: article.tex
\documentclass[accepted,mathfont=cm]{uai2021pre} % after acceptance, for a revised
\usepackage[american]{babel}
\usepackage{natbib} % has a nice set of citation styles and commands
\usepackage{mathtools} % amsmath with fixes and additions
\usepackage{booktabs} % commands to create good-looking tables
\usepackage{tikz} % nice language for creating drawings and diagrams

\usepackage{graphicx}
\usepackage{booktabs} % for professional tables
\usepackage{algorithm}
\usepackage{subcaption}
\usepackage[noend]{algpseudocode}
\usepackage{xcolor}
\usepackage{stfloats}

\usepackage{url}            % simple URL typesetting
\usepackage{graphicx}
\usepackage{amsfonts,amsmath,amssymb,mathtools}       % blackboard math symbols
\usepackage{tikz,pgfplots}
\pgfplotsset{compat=newest}
\usepgfplotslibrary{groupplots,dateplot}
\usepgfplotslibrary{external}
\graphicspath{{./fig/}}

\newlength{\figurewidth}
\newcommand{\N}{\mathcal{N}}

\renewcommand{\v}[1]{\boldsymbol{#1}} %vector
\newcommand{\m}[1]{\boldsymbol{#1}} %matrix
\newcommand{\op}[1]{\operatorname{#1}} %matrix
\newcommand{\g}{\mid}
\newcommand{\R}{\mathbb{R}}
\newcommand{\Trans}{^{\intercal}}

\definecolor{grayC}{HTML}{8B8B8B}
\definecolor{blueC}{HTML}{008FD5}
\definecolor{greenC}{HTML}{6D904F}
\definecolor{purpleC}{HTML}{810F7C}

\definecolor{lowcolor}{rgb}{0.987053, 0.991438, 0.749504}
\definecolor{midcolor}{rgb}{0.716387, 0.214982, 0.47529}
\definecolor{highcolor}{rgb}{0.002258, 0.001295, 0.018331}

\definecolor{l4lowcolor}{rgb}{0.031373, 0.1882353, 0.419608}
\definecolor{l4highcolor}{rgb}{0.4, 0.4, 0.4}
\newcommand{\figline}[1]{\raisebox{2pt}{\tikz{ \draw[-,#1,solid,line width = 1.3pt](0,0) -- (4mm,0);}}}

\title{A Probabilistically Motivated Learning Rate Adaptation\\ for Stochastic Optimization}

\author[1,2]{\href{mailto:Filip de Roos <filip.de.roos@tuebingen,mpg.de>?Subject=A Probabilistically Motivated Learning Rate Adaptation}{Filip de Roos}{}}
\author[3]{Carl Jidling}
\author[4]{Adrian Wills}
\author[3]{Thomas B. Sch\"{o}n}
\author[1,2]{Philipp Hennig}
\affil[1]{%
    Department of Computer Science\\
    University of T\"{u}bingen, Germany
}
\affil[2]{%
    Max Planck Institute for Intelligent Systems\\
    T\"{u}bingen, Germany
}
\affil[3]{%
    Department of Information Technology\\ 
    Uppsala University, Sweden
}
\affil[4]{School of Engineering\\ 
  University of Newcastle, Australia}

\begin{document}
\maketitle

\begin{abstract}
Machine learning practitioners invest significant manual and computational resources in finding suitable learning rates for optimization algorithms. We provide a probabilistic motivation, in terms of Gaussian inference, for popular stochastic first-order methods. As an important special case, it recovers the Polyak step with a general metric. The inference allows us to relate the learning rate to a dimensionless quantity that can be automatically adapted during training by a control algorithm. The resulting meta-algorithm is shown to adapt learning rates in a robust manner across a large range of initial values when applied to deep learning benchmark problems.
\end{abstract}

\section{INTRODUCTION} % (fold)
  \label{sec:introduction}
% textwidth: \printinunitsof{in}\prntlen{\textwidth}
% linewidth: \printinunitsof{in}\prntlen{\linewidth}
  % Font:\showthe\font
  % \scriptsize{\showthe\font}
  % \showthe\font
  Empirical risk minimization, including in particular deep learning, requires optimization of an objective $f$ that is the sum of individual losses $\ell$ over elements $\v{x}_d$ of the dataset $\mathcal{D}$ with $|\mathcal{D}|=D$ as
  \begin{equation}
  f(\v{\theta} )= \frac{1}{D}\sum_{d=1} ^{D} \ell(\v{\theta}, \v{x}_d) + \mathcal{R}(\v{\theta}) ,
  \end{equation}
  where $\v{\theta}\in \R^N$ denotes the parameter to be optimized and $\mathcal{R}$ is an additional regularization term.
  It is standard practice to sub-sample the dataset into batches $\mathcal{B}\subset \mathcal{D}$ with $|\mathcal{B}|=B$ when evaluating the loss $f$ and its gradient, yielding a stochastic, noisy observation 
  \begin{equation}
  \ell_B(\v{\theta}) = \frac{1}{B}\sum\limits_{b=1} ^{B} \ell(\v{\theta}, \v{x}_b) + \mathcal{R}(\v{\theta}). 
  \label{eq:batch_loss}
  \end{equation}
  If the elements of the batch are drawn i.i.d. and $1\ll B\ll D$, then by the Central Limit Theorem, $\ell_B$ is approximately Gaussian distributed around the true function value
  \begin{equation}
  \label{eq:ell_distribution}
  p(\ell_B) = \N(\ell_B;f(\v{\theta}),R),
  \end{equation}
  with variance $R$ that scales as $\mathcal{O}(1/B)$. 
  While stochasticity drastically reduces computational cost and may have beneficial side-effects like improved generalization \citep{hardt2016train,wu2020noisy}, it also complicates parameter tuning. 
  In contrast to classic numerical optimization routines, variants of stochastic gradient-descent (SGD) %\cite{robbins1951stochastic,kiefer1952} 
  expose free parameters to the user. 
  Chief among them is the scalar learning rate $\eta$ that determines the step size taken in the direction $\v{v}_i$ as 
  \begin{equation}
    \v{\theta}_{i+1} = \v{\theta}_{i} - \eta \cdot \v{v}_i, 
    \label{eq:standard_update}
  \end{equation}
  with $\v{v}_i$ chosen iteratively by the optimization routine (in the case of vanilla SGD, $\v{v}_i=\nabla \ell_B (\v{\theta}_i)$). 
  The learning rate constitutes a crude approximation to local curvature and crucially affects the convergence of the training, and by extension the performance of the model. 
  Its optimal value depends on the network architecture, the dataset, and the optimization method. %; these aspects combined yields a broad range of possible learning rates. 
  Although various semi-automated and fully automated routines have been proposed to tune learning rates \citep{baydin2017hyper,mahsereci2017probabilistic,vaswani2019painless}, practitioners still largely rely on a manual process of repeated training runs, causing significant use of computational resources~\citep{asi2019importance}.

  \subsection{Contributions}
  In this work, we describe a probabilistic inference scheme that can be used as an \textit{add-on to existing} first-order optimization methods (Section~\ref{sec:method}).  
  The procedure explicitly models observation noise caused by data subsampling which in the noise-free limit recovers a generalization of the Polyak (\citeyear{polyak87optimization}) step for parameter updates. In Section \ref{sub:choice_of_covariance} we show how various well-known optimization methods can be included in the inference and pave the way for the identification of new ones. 
  There are several parameters associated with the inference procedure. We analyze them in detail and based on the findings arrive at a learning rate adaptation scheme (Section~\ref{sec:algorithm}).  
  It relies on a local quadratic model of the loss function, implicitly defined by the underlying optimization algorithm. The learning rate is adapted during training thereby reducing the need for outer-loop tuning.
  %The key point of our model is that it associates the change of the objective function's value during a single update as an ``innovation'', a dimensionless ratio between a prediction residual and an expected value for the residual. 
  % With empirical estimators for the terms in the innovation (Section~\ref{sub:parameters}), we design a simple control algorithm that keeps the innovation close to unit value by adapting the step size during training, yielding a step-size control.
  We empirically show that the proposed update rule is robust w.r.t. the learning rate, leading to stable convergence for a range of popular optimization algorithms across common deep learning benchmarks.

\section{METHOD}
\label{sec:method}
To set the scene, consider a refactored second-order Taylor expansion of the function $f(\v{\theta}):\R^{N}\rightarrow \R$ in a vector $\v{d}$ around the current location $\v{\theta}_i$, using the (ground-truth, full-batch) gradient $\nabla f_i\triangleq\nabla f(\v{\theta})\g_{\v{\theta}=\v{\theta}_i}$ and Hessian $\m{B}_i\in\R^{N\times N}$ with $[\m{B}_i]_{mn}\triangleq\frac{\partial^2 f(\v{\theta})}{\partial \v{\theta}_m\partial \v{\theta}_n}\g_{\v{\theta}=\v{\theta}_i}$. 
Assuming the Hessian is invertible, we write this local approximation as
\begin{equation}
\begin{split}
\bar{f}_i(\v{\theta}_i+ \v{d}) 
&=\underbrace{\frac{1}{2}(\v{d}+\m{B}_i^{-1}\nabla f_i )\Trans \m{B}_i(\v{d}+\m{B}_i^{-1}\nabla f_i )}_{\phi_{\m{B}}(\v{d})} \\ 
&+   \underbrace{f(\v{\theta}_i) -\frac{1}{2}\nabla f_i\Trans \m{B}_i^{-1}\nabla f_i }_{f^{*}}. 
\end{split} 
\label{eq:taylor}
\end{equation}
When the Hessian is positive definite, the minimum value of this quadratic approximation $f^{*}$ is attained at the well-known Newton update $\v{d}_i^\star=-\m{B}_i^{-1}\nabla f_i$ which occurs at the point $\v{\theta}_i^\star= \v{\theta}_i -\m{B}_i^{-1}\nabla f_i$.
Because computing this update is computationally costly, large parts of classic convex optimization (in particular, conjugate gradients and quasi-Newton methods \citep[e.g.][\textsection 5 \& 6, respectively]{nocedal2006numerical}) are concerned with efficient estimation of $\v{d}_i^\star$ from a sequence of observed gradients. 
Big-data machine learning adds a new challenge to this setting, for which these classic methods are ill-equipped: significant sub-sampling noise on the gradient and (if it is computed) the Hessian.

\subsection{Probabilistic Model}
\label{sub:probabilistic_model}

We phrase the task of locating (inferring) the minimizer $\v{\theta}_i^\star \in \mathbb{R}^N$ of the local quadratic model given in Eq.~\eqref{eq:taylor} at iteration $i$ based on noisy observations of the cost $\ell_B(\v{\theta}_i)$ from Eq.~\eqref{eq:batch_loss} as a probabilistic inference problem: 
We model $\v{\theta}_i^\star$ as a random variable, denoted $\widehat{\v{\theta}}_i^\star$, and compute the posterior distribution of $\widehat{\v{\theta}}_i^\star$ conditioned on $\ell_B(\v{\theta}_i)$ via Bayes rule
\begin{align}
p(\widehat{\v{\theta}}_i^\star \mid \ell_B(\v{\theta}_i)) =\frac{p(\ell_B(\v{\theta}_i) \mid \widehat{\v{\theta}}_i^\star)\, p(\widehat{\v{\theta}}_i^\star)}{p(\ell_B(\v{\theta}_i))}.
\label{eq:bayes}
\end{align}
The prior $p(\widehat{\v{\theta}}_i^\star)$ is taken to be Gaussian and centered around the current parameter value $\v{\theta}_i$: 
\begin{align}
p(\widehat{\v{\theta}}_i^\star) = \mathcal{N}(\widehat{\v{\theta}}_i^\star;\, \v{\theta}_i,\, \m{W}_i),
\label{eq:prior}
\end{align}
where $\m{W}_i\in\R^{N\times N}$ is an arbitrary symmetric positive definite covariance
matrix, which is discussed further in Section~\ref{sub:choice_of_covariance}.
To develop the likelihood $p(\ell_B(\v{\theta}_i) \mid \widehat{\v{\theta}}_i^\star)$, we start by rewriting Eq.~\eqref{eq:taylor} in terms of ${\v{\theta}}_i^\star$ as
\begin{equation}
  \begin{split}
\bar{f}_i(\v{\theta}_i)|_{\v{d}=0}
&= \bar{f}_i(\v{\theta}_i - \v{d}_i^\star + \v{d}_i^\star) \\
&= \bar{f}_i(\v{\theta}_i + \v{\theta}_i - \v{\theta}_i^\star + \v{d}_i^\star).
\end{split}
\end{equation}
Inserting this statement in Eq.~\eqref{eq:taylor} and recalling that $\v{\theta}_i^\star -\v{\theta}_i=\v{d}_i^\star=-\m{B}_i^{-1}\nabla f_i$, we obtain
\begin{equation}
\begin{split}
  \bar{f}_i(\v{\theta}_i) &= \frac{1}{2}(\v{\theta}_i-\v{\theta}_i^\star)\Trans \m{B}_i(\v{\theta}_i-\v{\theta}_i^\star) +f^*, \\ \text{ and } \quad
  \nabla \bar{f}_i &= \m{B}_i(\v{\theta}_i-\v{\theta}_i^\star).
\end{split}
\label{eq:surrogate}
\end{equation}
Since the quadratic approximation matches the true function and its gradient at $\v{\theta}_i$, we note that
$
f(\v{\theta}_i)
=\bar{f}_i(\v{\theta}_i)
%= \frac{1}{2}(\v{\theta}_i-\v{\theta}_*)\Trans \m{B}_i(\v{\theta}_i-\v{\theta}_*) +c
= \frac{1}{2}\nabla \bar{f}_i\Trans (\v{\theta}_i-\v{\theta}_i^\star) + f^*$.

This finding motivates us to express the noisy observation $\ell_B(\v{\theta}_i)$ in the \textit{probabilistic} model as the following linear projection
\begin{equation}
  \begin{split}
    \ell_B(\v{\theta}_i) &= \frac{1}{2}\v{g}_i\Trans (\v{\theta}_i-\widehat{\v{\theta}}_i^\star) + f^* +\epsilon_i, \\ 
     \epsilon_i &\sim\mathcal{N}(0,R_i),
  \end{split}
  \label{eq:lb_theta_star}
\end{equation}
or equivalently by the likelihood
\begin{equation}
  p(\ell_B(\v{\theta}_i) \g \widehat{\v{\theta}}_i^\star) 
  = \mathcal{N}\left(\ell_B(\v{\theta}_i); 
  \frac{1}{2}\v{g}_i\Trans (\v{\theta}_i - \widehat{\v{\theta}}_i^\star) + f^*, R_i \right).
  \label{eq:likelihood}
\end{equation}
Here we use $\v{g}_i$ to denote a the gradient of the loss over the mini-batch and $R_i$ is the observation noise due to subsampling.
Under the Gaussian prior \eqref{eq:prior} and linear likelihood \eqref{eq:likelihood} there is a closed-form expression\footnote{For $p(\v{x})=\mathcal{N}(\v{x};\v{\mu},\m{\Sigma})$ with $\v{x},\v{\mu}\in\R^N,\m{\Sigma}\in\R^{N\times N}$ and $p(y\g \v{x}) = \mathcal{N}(y;\m{A}\v{x} + b,R)$ with $\v{A}\in\R^{1\times N}$ and $b\in\R,R\in\R_+$, it holds that
$
  p(\v{x}\g y) = \mathcal{N}\big(\v{x};\v{\mu} + \frac{\m{\Sigma}\m{A}\Trans (y - \m{A}\v{\mu} - b)}{\m{A\Sigma A}\Trans + R}, \m{\Sigma}-\frac{\m{\Sigma}\m{A}\m{A}\Trans\m\Sigma}{\m{A\Sigma A}\Trans + R}\big). 
$
} for the posterior, stated in Eq. \eqref{eq:bayes}. The posterior mean will serve as our next estimate, $\v{\theta}_{i+1}$, and is given by
\begin{equation}
\label{eq:mean_update}
  \v{\theta}_{i+1} = \v{\theta}_i - \m{W}_i \v{g}_i \frac{ 2(\ell_B(\v{\theta}_i) - f^*)}{\v{g}_i\Trans \m{W}_i \v{g}_i + R_i}.
\end{equation}
This parameter update is of the same form as the generic update in Eq.~\eqref{eq:standard_update} \emph{if} $\v{v}_i=\m{W}_i \v{g}_i$. In the next section we will clarify how this update corresponds to popular first-order algorithms and later look into the different parameters that are required for the inference.

\subsection{Choice of covariance} % (fold)
\label{sub:choice_of_covariance}
The update in Eq.~\eqref{eq:mean_update} leaves the prior covariance matrix $\m{W}_i$ as a free parameter. 
To be a valid covariance, it must be symmetric and positive definite. 
For the batch gradient $\v{g}_i$ (i.e. $\nabla \ell_B$), a step in direction $\v{v}_i=\m{W}_i \v{g}_i$ is a descent direction (on the batch), because $-\v{g}_i\Trans \v{v}_i<0$. 
To clarify the connection to optimization algorithms with a provided learning rate we will refer to $\m{W}_i$ as $\eta \cdot \m{W}_i$, with $\eta\in \R^+$.
For different choices of $\m{W}_i$, Eq.~\eqref{eq:mean_update} can be seen as a generalization of several existing optimization algorithms. 

\paragraph{SGD} % (fold)
\label{par:sgd}
The most straightforward connection is to 
SGD~\citep{robbins1951stochastic}: If we consider $\eta\cdot \m{W}_i=\eta\cdot \m{I}$, Eq.~\eqref{eq:mean_update} becomes
\begin{equation}
  \v{\theta}_{i+1}=\v{\theta}_i - \eta \cdot \v{g}_i \frac{2(\ell_B(\v{\theta}_i)-f^*)}{ \eta  \|\v{g}_i\|^2 + R_i} 
  \label{eq:SGD}.
\end{equation}
For $R_i=0$ this update recovers the Polyak step \citep{polyak87optimization,loizou2020stochasticpolyak} if $f^*$ is known.
A correctly identified $f^*$ in the probabilistic argument above motivates a learning-rate adaptation for SGD. 
As the optimizer approaches the optimum, $\ell_B \gtrsim f^*$, the rate goes towards zero. If instead the fraction above is constant across iterations, we recover the standard update for SGD with fixed learning rate.

\begin{table}
% \vspace{-10pt}
\centering
\begin{tabular}{lc}
\toprule
 Optimizer  & $\eta \cdot \m{W}_i$ \\
 \midrule 
 SGD      & $\eta \m{I}$  \\
 Adagrad  & $\eta \left( \m{G}_{i-1} + \op{diag}(\v{g}_i \v{g}_i\Trans ) \right)^{-1/2}$ \\
 RMSprop  & $\eta \left( \alpha\m{G}_{i-1} + (1-\alpha)\op{diag}(\v{g}_i \v{g}_i\Trans ) \right)^{-1/2}$ \\
 Momentum & $\eta  \left( \m{I} + \tilde{\beta}  \v{m}_{i-1}\v{m}\Trans_{i-1} \right)$ \\
 Adam     & $\eta\left( (1-\beta_1)\m{V}_i + \tilde{\beta}_1  \m{V}_i\v{m}_{i-1}\v{m}_{i-1}\Trans\m{V}_i \right)  $ \\
\bottomrule
\end{tabular}
\caption{Covariance matrices used for popular optimization algorithms. Each consists of a diagonal matrix and the last two have an additional rank one update with a modified scaling which we elaborate on in Section \ref{sub:accelerated_gradient}. $\m{G}_i$ for Adagrad and RMSprop are recursively defined as the quantity inside the parenthesis starting from $\m{G}_0 = 0$. The diagonal matrix $\m{V}_i$ for Adam is the $\m{G}_i^{-1/2}$ of RMSprop scaled with additional bias correction.} 
\label{tab:optimizers}
\end{table}  

\paragraph{General Optimizer} % (fold)
\label{par:general}
There has been a surge of stochastic first-order optimization algorithms to tackle the requirements of machine learning and deep learning in particular, many of which employ an element-wise scaling to the batch gradient $\v{g}_i$. 
These element-wise updates correspond to a diagonal matrix $\m{W}_i$ in Eq.~\eqref{eq:mean_update} which is the definition of an axis-aligned Gaussian distribution for the prior. 
Tab.~\ref{tab:optimizers} summarizes a few popular diagonal first-order optimizers in the probabilistic view together with a novel interpretation of acceleration in the form of momentum.

The inference is not limited to a diagonal $\m{W}_i$, it is just a computationally efficient approach to speed up first-order methods. Several higher order optimization methods can be included as well. In particular, if $\m{W}_i^{-1}$ is chosen as a curvature matrix, e.g.,~the Hessian, Gauss-Newton, or the Fisher information matrix, then the inference amounts to an adaptive version of Newton, Gauss-Newton or Natural Gradient Descent, respectively.

In addition to the posterior mean one could also consider the posterior covariance on $\widehat{\v{\theta}}_i^\star$. It could be propagated through optimization using a predictive Chapman-Kolmogorov equation. This could be realized as a Kalman filter prediction step, but would introduce additional empirical parameters and cost. 
We omit the covariance update to avoid confusion and instead focus on the useful connection to first-order optimization algorithms.

\subsection{Accelerated gradient updates} % (fold)
\label{sub:accelerated_gradient}
Several popular optimization algorithms employ an exponential averaging of gradients in the form of momentum \citep{polyak87optimization,Sutskever2013,Kingma2014_Adam} to speed up the training. Such methods can be included in the probabilistic inference in two ways.
 The first is to interpret the momentum term as another estimate of the \emph{true} gradient instead of the batch gradient, essentially replacing $\v{g}_i$ with $\v{m}_i$ in the derivations of Section.~\ref{sec:method}.
 The second way that we opted for is to retain the batch gradient but adjust the covariance with a rank one update.

The Pytorch implementation of SGD with momentum uses the following update:
\begin{align*}
 \v{m}_i &= \beta \cdot \v{m}_{i-1} + \v{g}_i, \\
 \v{\theta}_i &= \v{\theta}_{i-1} - \eta \cdot \v{m}_i,
\end{align*}
where $\beta$ is a hyperparameter controlling the influence of previous gradients.
This update is identified in the probabilistic model with
\begin{equation*}
  \eta \cdot \m{W}_i \v{g}_i = \eta \cdot \left( \m{I} + \frac{\beta}{\left<\v{m}_{i-1},\v{g}_i \right>} \cdot \v{m}_{i-1}\v{m}_{i-1}\Trans  \right) \v{g}_i.
  \end{equation*}

In a similar manner it is possible to express the update of Adam with the diagonal matrix 
\begin{align*}
  \m{V}_i  &= \gamma_i \cdot \left( \beta_2 \cdot \m{G}_{i-1} +  (1-\beta_2)\cdot \op{diag}(\v{g}_i \v{g}_i\Trans ) \right)^{-1/2},\\
  \gamma_i &= \sqrt{(1-\beta_2^i)}/(1-\beta_1^i),
\end{align*}
and the exponential average $\v{m}_i = \beta_1 \v{m}_{i-1} + (1-\beta_1)\v{g}_i$ as
\begin{align*}
\eta\cdot \m{W}_i \v{g}_i &=\eta \cdot \left( (1-\beta_1)\m{V}_i + \tilde{\beta}_1 \m{V}_i \v{m}_{i-1} \v{m}_{i-1}\Trans \m{V}_i  \right) \v{g}_i,\\
   \tilde{\beta_1} &=\frac{\beta_1}{\left< \m{V}_i \v{m}_{i-1}, \v{g}_i \right>}.
\end{align*}
Neither the Momentum nor Adam update is positive definite, per definition, in this form due to the scalar parameter in front of the rank-1 update. For SGD with momentum this scalar value will be negative if $\left< \v{m}_{i-1}, \v{g}_i \right> <0$ and violates the positive definiteness if $\beta \left< \v{m}_{i-1}, \v{g}_i \right> < -\left< \v{g}_{i}, \v{g}_i \right>$. This occurs when $\v{m}_{i-1}$ is pointing significantly uphill and the situation is more likely to arise for high values of $\beta$. The conditions are analogous for Adam.

\section{ALGORITHM} % (fold)
\label{sec:algorithm}
\setlength{\figurewidth}{0.99\columnwidth}
\begin{figure}[ht]%{\figurewidth}
    \centering
    % \includegraphics[width=\figurewidth]{}
  % \vspace{-22pt}
    \input{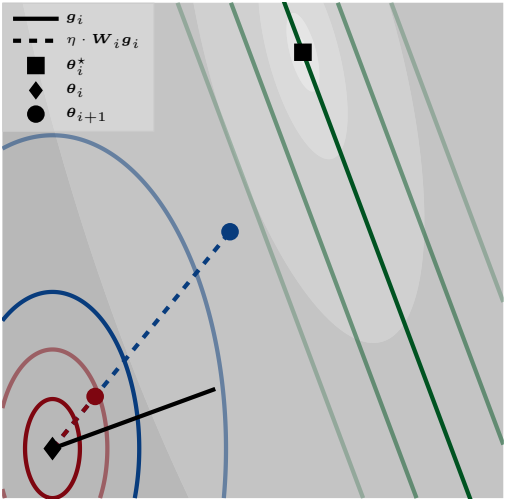}
    \caption{Illustration of the probabilistic inference scheme.
        The gray contour levels describe the local quadratic model $\bar{f}_i(\v{\theta})$. 
        The dark green line comprises all \textit{possible} values of $\widehat{\v{\theta}}_i^\star$ (orthogonal to $\v{g}_i$) that satisfy Eq. \eqref{eq:lb_theta_star} (in the noise-free case); 
        %$1/2 \cdot\bar{\v{g_i}}\Trans (\v{\theta}_i %-\v{\theta})+c=\bar{f}(\v{\theta}_i)$, 
        this includes the \textit{true} value $\v{\theta}_i^\star$. 
        The lighter green levels are multiples of the standard deviation $\sqrt{R_i}$ representing the uncertainty in Eq.~\eqref{eq:lb_theta_star}. 
        The red and blue ellipses illustrate two axis-aligned Gaussian distributions (Eq.~\eqref{eq:prior}) centered at $\v{\theta}_i$, corresponding to different scaling of the covariance $\m{W}_i$.
        The circles indicate the posterior mean ($\v{\theta}_{i+1}$ in Eq.~\eqref{eq:mean_update}) for each distribution. 
        The blue distribution has a larger variance relative to $R_i$ leading to a larger step towards the solution compared to the red. 
    }
    \label{fig:2d}
\end{figure}
The update in Eq.~\eqref{eq:mean_update} is the most general form of the provided inference scheme from which one can approximate or specialize the update depending on available information and problem.
While the structure of $\eta\cdot \m{W}_i$ should be seen as a design choice addressing an algorithm within a larger family, the remaining parameters are typically problem-dependent and play an important role in the convergence of the underlying algorithm.
This section addresses this issue and constructs a simple learning rate adaptation scheme. 
An advantage of the probabilistic derivation is that it offers an interpretation of these parameters, which can be used to construct empirical estimators. 
Pseudo-code for our method that is implemented as a wrapper applicable to valid optimization algorithms is provided in Alg.~\ref{alg:pseudo}.

\subsection{Parameters} % (fold)
\label{sub:parameters}
Depending on the problem, the update scheme requires estimation of up to three parameters: The scale of the prior variance $\eta$, the lower bound $f^*$ and the observation variance $R_i$. With the probabilistic motivation developed in Section~\ref{sec:method}, it is possible to link these parameters to function values and hence estimate them at runtime. These three parameters have an interesting interplay---the same update of $\v{\theta}_{i+1}$ (Eq.~\ref{eq:mean_update}) can arise from different constellations of these three numbers. In traditional optimization, uncertainty of the observation is typically not considered, which is why $R_i$ does not appear in the standard Polyak step. 

\paragraph*{Uncertainty $R_i$}
The most straightforward parameter to estimate is the observation variance $R_i$ in Eq.~\eqref{eq:mean_update}. In the case of minibatching it is possible to estimate the uncertainty of the full loss, as outlined via the CLT argument in Section~\ref{sec:introduction}. For general functions the situation is more complicated and since the inclusion of $R_i$ corresponds to a reduction in step length, the same effect can be achieved by omitting $R_i$ and instead decrease the difference in the numerator.

\paragraph*{Scale $\eta$}
The second influential parameter is $\eta$, a scalar multiplication of the prior covariance corresponding to the learning rate of the optimizer. For the general case of Eq.~\eqref{eq:mean_update} with $R_i > 0$, the ratio between $R_i$ and $\eta \cdot \v{g}_i\Trans \m{W}_i \v{g}_i$ becomes important for the overall step length and hence the convergence of the algorithm. This behavior is visualized in Fig.~\ref{fig:2d} where the variance of the red distribution is low compared to $R_i$, leading to small updates. A result of this is that the initial learning rate can be arbitrarily ill-calibrated with regards to the noise variance. It also suggests that updating the scale $\eta$ during the optimization, once more parameters can be estimated, could lead to improved performance. If instead $R_i=0$ and $f^{*}$ in Eq.~\eqref{eq:mean_update} is known, then the update will not depend on the scale $\eta$.

\paragraph*{Lower bound $f^*$}
The final parameter $f^*$ is also the most important in terms of performance and stability. For a known $f^*$ and $R_i=0$, the Polyak step achieves a linear convergence rate \emph{towards $f^*$}, independent of the Lipschitz constant which normally bounds the convergence of first-order optimization algorithms  \citep{polyak87optimization,vaswani2020armijols}. If $f^*$ is not known and set too large, then the optimizer will converge to parameters for which $f(\v{\theta})=f^*$ and not the actual minimum. If instead $f^*$ is not known and estimated below the minimum, then the Polyak step will try to reach function values below the minimum. This is problematic in flat regions, resulting in steps that are too large which can undo a lot of progress (see Appendix~\ref{A:probabilistic_model}). To combat this behavior, a maximum step length is often introduced as a problem dependent hyperparameter \citep{berrada2019alig,loizou2020stochasticpolyak}. The same authors showed that for Machine learning problems with overparameterized models that fulfill \textit{interpolation} (models which can achieve a training loss close to 0 for all training samples simultaneously) it is possible to use $f^*=0$ as a lower bound for the empirical risk minimization for fast convergence. 

\begin{algorithm}[h]
\begin{algorithmic}[1]
\Procedure{LR-Adapt}{$\v{\theta}_1$, $\eta$, $F(\v{\theta})$, $G(\v{\theta})$, $W(\v{g})$, $f^*$, $R_i=0$}

\For{$i = 1 \dots$}
\State $f_i \gets F(\v{\theta}_i)$ \Comment{Evaluate function }
% \State $R_i \gets \op{Var}[F(\v{\theta}_i)]$ \Comment{Variance scaled with batch-size}
\State $\v{g}_i\gets G(\v{\theta}_i)$ \Comment{Obtain gradient}
\State $\v{v}_i \gets \eta \cdot W(\v{g}_i)$ \Comment{$\v{v}_i =\eta \cdot \m{W}_i \v{g}_i$ }
\State $\phi_i \gets \v{g}_i\Trans \v{v}_i$\\ %\Comment{Expected decrease under quadratic model times 2}\\

\If{$f^*$ is available}
  \State $\Delta f\gets f_i - f^*$ %$\Delta_{i}\gets \phi_i/2$ \Comment{Use provided step length for first step}
\Else 
    \State $\Delta f \gets f_i - \phi_i/2$ %\Comment{Use expected decrease of implied quadratic}
\EndIf\\

\State $\v{\theta}_{i+1}\gets \v{\theta}_{i} - \v{v}_i  \cdot 2\cdot \frac{\Delta f}{\phi_i+R_i}$ \\

\State $f_+ = F(\v{\theta}_{i+1})$ \Comment{Re-evaluate same batch} \\

% \State $err \gets (f_{t-1}-f_{i}) -\phi_i$ \Comment{Error between observation and model}
\If{$(f_i - f_+)/(\phi/2)>4/3$}
\State $\eta \gets 1.2\cdot \eta$ \Comment{$\eta$ too small $\alpha_\uparrow$}
\ElsIf{$(f_i - f_+)/(\phi/2)<3/4$}
\State $\eta \gets 1/2\cdot \eta$ \Comment{$\eta$ too large $\alpha_\downarrow$} \label{alg:scale_down}
\EndIf

% \State $f_{i-1}\gets f_{i}$ \Comment{Save function value for next iteration}
% \EndWhile\label{euclidendwhile}
\EndFor
% \State \textbf{return} $b$\Comment{The gcd is b}
\EndProcedure
\end{algorithmic}
\caption{\label{alg:pseudo}Step size adaptation for a provided function evaluator $F(\v{\theta})$, gradient estimator $G(\v{\theta})$, and search direction selector $W(\v{g})$ defined by the underlying optimizer.}
\end{algorithm}

\subsection{Implementation}
The previous section showed that the relatively simple update of Eq.~\eqref{eq:mean_update} requires careful consideration, see Appendix~\ref{A:probabilistic_model} for an example. Here we will outline the steps taken in Alg.~\ref{alg:pseudo} to address this. Once a valid optimizer has been selected all the quantities up to and including line 7 are accounted for. A step of our algorithm then requests two parameters: a lower bound $f^*$ and an uncertainty estimate $R_i$ (defaults to 0).
In the absence of externally provided lower bound $f^*$ (default behavior), we still want the algorithm to adapt the learning rate during optimization in a useful manner. The approach that we use for this situation is to consider an \textit{implied} quadratic, which given a function value $f(\v{\theta}_i)$, gradient $\v{g}_i$ and a spd matrix $\eta_i \cdot \m{W}_i$ constructs a local surrogate
\begin{equation}
\begin{split}
  \phi_{\m{W}_i}(\v{d}) = \phi_{\text{min}} &+\\
   \frac{1}{2}\left( \v{d} + \eta_i \m{W}_i\v{g}_i \right)\Trans&(\eta_i \m{W}_i)^{-1}\left(  \v{d} + \eta_i \m{W}_i\v{g}_i \right). 
\end{split}  
\end{equation}
The parameters are chosen such that $\phi_{\m{W}_i}(\v{0})=f(\v{\theta})$ and $\nabla_{\v{d}}\phi|_{\v{d}=0}=\v{g}_i$. The minimum of this surrogate occurs at the step $\v{d}= -\eta_i \m{W}_i \v{g}_i$ corresponding to a decrease of $\phi(0)-\phi(-\eta_i \m{W}_i \v{g}_i) = \eta_i\cdot \v{g}_i\Trans \m{W}_i \v{g}_i /2$. This lower bound in Eq.~\eqref{eq:mean_update} amounts to a standard step of the underlying optimization algorithm if $R_i=0$, but with an additional advantage. By re-evaluating the function at the new parameter, one can adapt the scale of the covariance/learning rate so $\phi_{i} - \phi_{i+1} = \eta_i \cdot \v{g}_i\Trans \m{W}_i \v{g}_i /2 \approx f(\v{\theta}_i) - f(\v{\theta}_{i+1})$. 
The decision of which lower bound to use occurs in lines 8 to 12 of Alg.~\ref{alg:pseudo} and is followed by the update in Eq.~\eqref{eq:mean_update}.

In line 15 we re-evaluate the function (same batch) and then compare the ratio of observed and expected decrease. A ratio of 1 corresponds to a perfect match of curvature between the real function and the estimated quadratic in direction $\v{v}_i$. If the ratio deviates from 1, we adapt the learning rate by multiplicatively increasing or decreasing $\eta$ with $\alpha_{\uparrow}=1.2$ and $\alpha_{\downarrow}=1/2$, inspired by the updates of Rprop \citep{riedmiller1992rprop}. Similar ideas are also employed in trust-region methods \citep[Ch. 4]{nocedal2006numerical} to adapt the size of the trust region. One could simply choose a new $\eta$ that makes the ratio 1 but this made the algorithm sensitive to outliers. Instead we apply the updates iteratively to guard against outliers which frequently occur due to the stochasticity of the problems considered. 
The asymmetry of the update is to penalize steps that are too large since these can be critical to the optimization. We allow a bit of deviation from the optimal value of the ratio to account for stochasticity of the gradients.   
%The iterative update is applied to guard against outliers which frequently occur due to the stochasticity of the problems considered. By re-evaluating the same function (mini-batch) we get a \emph{noise-free} ($R_i=0$) evaluation of \emph{this} function, and thus a stochastic learning rate estimate. 
As the learning rate is adapted for each mini-batch it approaches a value that is suitable for the full-batch function.

% \begin{figure}[ht]%
%   % \definecolor{darkred}{rgb}{0.7, 0.0, 0.0}
%   % \definecolor{darkblue}{rgb}{0.0, 0.0, 0.7}
%   % "#001C7F", "#B1400D", "#12711C", "#8C0800", "#591E71",
%             % "#592F0D", "#A23582", "#3C3C3C", "#B8850A", "#006374"
%   \definecolor{dred}{RGB}{160,0,0} 
%   \definecolor{dblu}{RGB}{0,0,130}
%   \centering%
%   \begin{tikzpicture}[scale=5]
%   \draw[->] (-0.1,0) -- (0.9,0);% node[right] {$x$};
%   \draw[->] (0,-0.1) -- (0,0.9);% node[above] {$y$};
%   \draw[scale=1,domain=0.1:0.9,smooth,very thick,variable=\x,dblu] plot ({\x},{4*(\x-0.5)*(\x-0.5)+0.2});
%   \draw[scale=1,domain=0.05:0.6,smooth,very thick,variable=\x,dred] plot ({\x},{-1.6*\x+0.92});
%   \draw[-] (0.1586,0) -- (0.1586,0.02) node[below=2mm] {$\v{\theta}_i$};
%   \draw[-] (0.4412,0) -- (0.4412,0.02) node[below=2mm] {$\v{\theta}_i-\alpha_i \v{p}_i$};
%   \draw[-] (-0.06,0.667) -- (0.02,0.667) node[left=4mm] {$f(\v{\theta}_i)$};
%   \draw[-] (-0.06,0.2) -- (0.02,0.2) node[left=4mm] {$c$};
%   \draw[-] (-0.04,0.2) -- node[left=0mm] {$\phi(\v{\theta}_i)$} (-0.04,0.667);
%   \end{tikzpicture}
%   \caption{Illustration of how a step-size $\alpha_i$ in direction $\v{p}_i$ is calculated for $\v{p}_i=\m{W}\v{g}_i$.}
%   \label{fig:phi}
% \end{figure}

\subsection{Computation} % (fold)
\label{sub:computation}
The overall cost of our algorithm is essentially one forward-pass of the batch loss more expensive than that of the underlying optimizer. This is due to the requirement of re-evaluating the batch loss before the next iteration. Apart from the re-evaluation there are two non-scalar operations involved in each step: finding the step direction $\v{v}_i = \eta \cdot \m{W}_i \v{g}_i$, and computing the inner product $\v{v}_i\Trans \v{g}_i$ (i.e.~$\eta\cdot \v{g}_i\Trans\m{W}_i \v{g}_i$) in Eq.~\eqref{eq:mean_update}. 
The former operation is handled by the underlying optimization algorithm which does not incur any additional computational cost since the optimizer must compute it regardless. The latter operation scales linearly with the number of parameters once $\v{v}_i$ is obtained, which is of similar complexity to the first-order optimizers in Tab.~\ref{tab:optimizers}. 
Optimizers in Pytorch usually compute the update $\v{v}_i$ per-parameter and apply the update immediately to save memory.
In order to keep the implementation as general as possible for the identified algorithms, we explicitly store the vector $\v{v}_i$ for the inner product. This storage can be avoided but would require implementing a new version of each optimizer. \\

\begin{figure}[ht]
\centering
\includegraphics{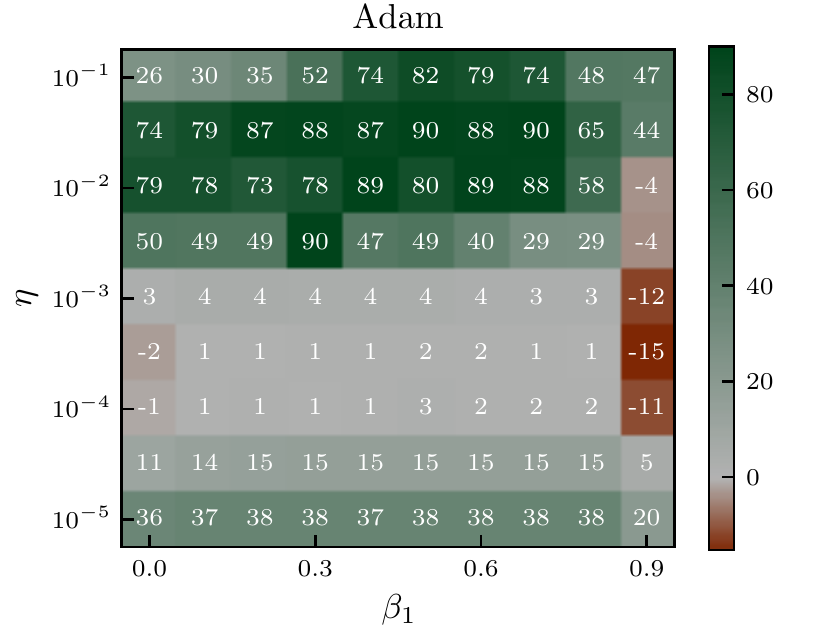}
\caption{Difference in achieved training accuracy between proposed adaptation and fixed learning rate version of Adam for different initial learning rates ($\eta$) and momentum ($\beta_1$). Green color means adaptation improved and red signifies worse accuracy after 50 epochs of training on an own implementation of the test problem cifar10\_3c3d. For each pair of parameters the experiment was repeated 3 times of which the best value is reported. Generally the adaptation leads to improvements except for large $\beta_1$, c.f. Section~\ref{sub:accelerated_gradient}. }
\label{fig:adam_heatmap}
\end{figure}

\section{EXPERIMENTS} % (fold)
\label{sec:experiments}
\begin{figure*}[ht]
\centering
\includegraphics{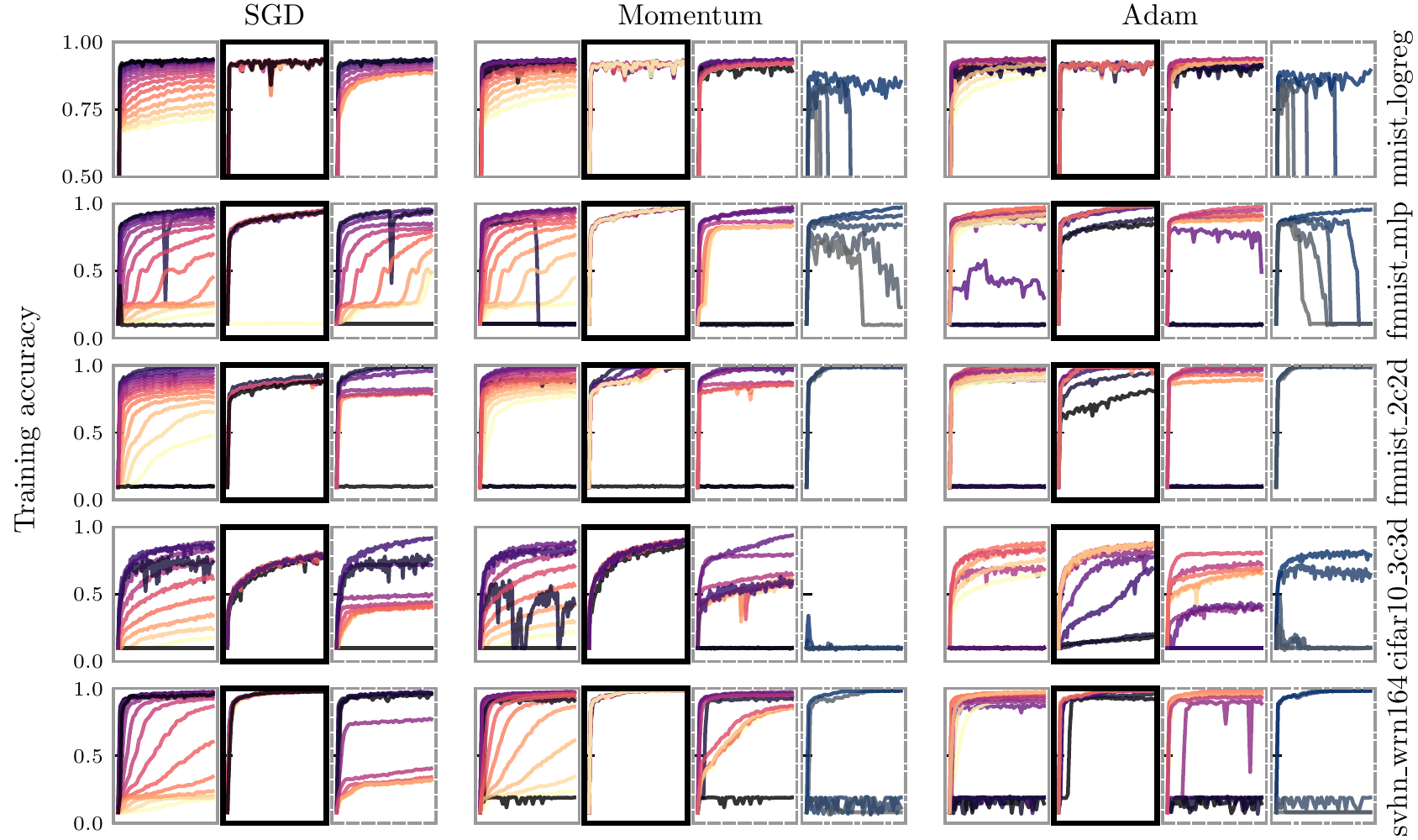}
\caption{\label{fig:deepobs}Training accuracy (higher is better) per epoch for different optimizers, learning rate adaptations and benchmark problems from DeepOBS. Each row shows the training accuracy for one test problem identified by a dataset and model description and each columngroup shows a family of optimizers. The leftmost graph in each group (thin gray border) has a fixed learning rate. Next to it (thick black border) is the proposed adaptation. A dashed border indicates results for Hypergradient descent and the dash-dotted show results for L4. Each graph contains experiments with initial learning rates in the range $10^{-5}$ (\protect\figline{lowcolor}) through $10^{-3}$ (\protect\figline{midcolor}) to $10^{0}$ (\protect\figline{highcolor}). In the case of L4 the learning is replaced with $\alpha_{L4}$ with values between 0.15 (\protect\figline{l4lowcolor}) and 0.25 (\protect\figline{l4highcolor}). In every problem each optimizer ran for 50 epochs except for cifar10\_3c3d which ran for 100 epochs. 
All hyperparameters were left at the default values except for the momentum term of the proposed adaptation which was set to 0.5 instead of default 0.9 for Momentum and Adam. The graphs under SGD show a typical example of how sensitive the performance of a model is to a fixed learning rate during training and how the adaptation avoids this problem.}
\end{figure*}

This section presents experimental results of relevant deep learning classification problems. We start by describing the different experiments and discuss the findings in the end. To ensure diversity in the problem set, reliable baseline comparisons and reproducible results, we made use of test problems provided by the DeepOBS benchmarking toolbox \citep{SchBalHen19deepobs}.\footnote{\label{foot:deepobs}\url{https://deepobs.github.io}}
We implemented our method in Pytorch \citep{paszke19pytorch} ver.~1.4 as a wrapper to the implemented optimizers listed in Tab.~\ref{tab:optimizers}. Across all experiments we used the default values of the parameters in Alg.~\ref{alg:pseudo}. The adaptation scheme (with no $f^*$ provided) is compared to a fixed learning rate, Hypergradient descent \citep{baydin2017hyper} and L4 \citep{rolinek2018l4} where applicable. 
In the absence of $f^*$ we found no significant difference in results by including an estimate of the noise variance $R_i$ or not, so it was left at 0.
To show the effect of using a poor learning rate and the efficacy of our adaptation, we ran each experiment and optimizer with initial learning rates in the range $10^{-5}$ to $1$. For L4 we varied $\alpha_{L4}$ which scales the numerator of Eq.\eqref{eq:mean_update} in the recommended range from the default value of 0.15 to 0.25. To better show the robustness of the proposed algorithm we report the results in terms of training accuracy since it is constrained to [0, 1]. All of the results can be seen in Fig.~\ref{fig:deepobs} and additional metrics and optimizers can be found in the supplementary material. Apart from the momentum term of Adam and SGD with momentum for our adaptation (further discussed below), all other hyperparameters were kept at the Pytorch default values.

\subsection{(F)MNIST} % (fold)
\label{sub:fmnist}
DeepOBS provides several test problems that are applicable to both the Fashion MNIST and standard MNIST dataset due to the identical data format. The first three rows of Fig.~\ref{fig:deepobs} show convergence results for models using logistic regression, a 4 layered multilayered perceptron and an artificial neural network with 2 convolutional layers followed by 2 dense layers for the classification.

\subsection{CIFAR-10} % (fold)
\label{sub:cifar_10}
The network used for the CIFAR-10 dataset \citep{Krizhevsky09learningmultiple} consists of 3 convolutional layers followed by 3 dense layers and $l_2$ regularization of $2\cdot 10^{-3}$. Each optimizer ran for 100 epochs as opposed to 50 for the other experiments due to the slower convergence. 
The same architecture was used to investigate how momentum affects our adaptation in deep learning. A typical result can be seen in Fig.~\ref{fig:adam_heatmap} illustrating a sweep across learning rates ($\eta$) and momentum ($\beta_1$) for Adam. The figure shows that the performance tend to deteriorate for large values of $\beta_1$. A similar sweep took place for SGD with momentum to settle on a momentum $\beta$ of 0.5 for both optimizers across all experiments. During the sweep we measured the average time it took to finish the training of one epoch and found that our algorithm required on average 41\% longer than the standard update.

\subsection{SVHN} % (fold)
\label{sub:svhn}

The SVHN dataset \citep{netzer2011svhn} contains more than $600\thinspace000$ images of house numbers seen from the street. One deepOBS architecture\textsuperscript{\ref{foot:deepobs}} used for this experiment is a \textit{wide resnet} \citep{Zagoruyko16WRN}, which is an extension of the \textit{deep resnet} \citep{He_resnet}. 
A key difference between the two architectures is that the wide resnet uses fewer and wider residual blocks, yielding improvements in training time, performance and number of parameters.
The network consists of 16 convolutional layers with a widening factor of 4, and we used a batch-size of 128 and $l_2$ regularization of $5\cdot 10^{-4}$.

\subsection{Discussion}
\label{sec:discussion}
The results presented in Fig.~\ref{fig:deepobs} show that the proposed learning rate adaptation is robust across a wide range of classification problems and optimizers. It reliably adjusts the learning rate which most times results in a final accuracy close to the best achieved accuracy on the task. The exception being SGD which in some cases falls behind due to the higher variance of $\eta\cdot \v{g}_i\Trans \m{W}_i \v{g}_i$ compared to other algorithms. 

Albeit the notable robustness, certain initializations of the learning rate are still too large for the optimization to converge, which is visible from the straggling dark/black lines in certain problems (an example is Adam for the fmnist tasks). In these cases the learning rates are several orders of magnitude larger than the optimal fixed learning rate and neither adaptation converges.   

Hypergradient descent often shows improvements over the corresponding fixed learning rate version but sometimes gets stuck for too small learning rates (see fmnist experiments) and it does not show the same agnosticism towards the initial learning rate. The update to the learning rate for Hypergradient descent is calculated from the inner product of two subsequent gradients and scaled with a small hyper learning rate. Since the size and architecture of the considered networks drastically vary between problems, the default value of the hyper learning rate is bound to be off for some architectures, requiring additional tuning. Our method instead updates the learning rate based on a dimensionless quantity making it less sensitive to variations in the network.

L4 estimates $f^*$ and uses a form of Polyak step in each iteration making each parameter update independent of the learning rate. The algorithm instead introduces additional hyperparameters which the authors empirically set for good performance. When L4 finished a training run without diverging it was usually among the fastest to reach a high accuracy, but the results in Fig.~\ref{fig:deepobs} show that the default parameter values would still require additional tuning depending on dataset and model making them less robust across problems.

Our implemented adaptation updates the learning rate for every batch throughout the training, leading to an overall computational cost on average $<50\%$ higher than that of the underlying optimizer. The additional cost stems from re-evaluating the loss on the same batch. 
% Preferably one would avoid the re-evaluation of the function (Line 15 of Alg.~\ref{alg:pseudo}) which would bring the cost of the algorithm to one additional $\mathcal{O}(N)$ operation per step, with $N$ the number or parameters of the model. 
A simple remedy to reduce the overhead is to not evaluate every batch or epoch, but every $2^i$th epoch for $i=0,1,...$. This allows significant adaptation in the beginning to get the scale right and less frequently during later stages of training, see Appendix~\ref{A:cifar100} for motivation. Overall, the additional cost of the re-evaluation is justified if it means that no additional runs are required to find a suitable learning rate.

One recurring observation from the experiments with the adaptation is that the smaller initial learning rates converge without exception, suggesting one could initialize the underlying optimizer with a learning rate of $10^{-4}-10^{-3}$ and let the adaptation accelerate.

\section{Related Work} % (fold)
\label{sec:relatedwork}
Stochastic gradient descent and its variants remain the workhorse for the stochastic optimization in deep learning, and big-data machine learning more generally. Several methods that improve the convergence over standard SGD by reducing the variance of the estimate \citep{Sutskever2013}, adapting the step-direction \citep{Duchi2011_adagrad,Zeiler2012_adadelta,Schaul2012_no_more_pesky_rate,Dauphin2015} or combinations thereof \citep{Kingma2014_Adam} have been proposed as substitutes \citep[see][for an overview]{Ruder2016_review}.
The learning rate is the single most important hyperparameter in these first-order optimization methods that are used in machine learning, with the model performance hinging on successful selection \citep{Goodfellow-et-al-2016}. Some recent ideas to reduce this influence are to include the learning rate as an additional parameter that can be optimized with backpropagation \citep[][cf.~results in Fig.~\ref{fig:deepobs}]{baydin2017hyper} or to train another model to predict the next step \citep{andrychowicz2016learn2learn}.

% A too large value can ruin the convergence, while a too small value makes the progress very slow and the optimizer risks getting completely stuck without further improvement \citep{Goodfellow-et-al-2016}. Some recent ideas for dealing with this are to include the learning rate as an additional parameter that can be optimized with backpropagation \citep[][cf.~results in Fig.~\ref{fig:deepobs}]{baydin2017hyper} or to train a model to predict the next step \citep{andrychowicz2016learn2learn}. 
% Speaking more generally, a large number of optimization routines for deep learning has recently been proposed, making it nearly impossible to compare to all of them in the confines of a conference paper. 
% In our experiments, we thus limit the comparison to the selection presented in table~\ref{tab:optimizers}.

In traditional optimization the learning rate problem is usually avoided by use of a line search routine, with new iterates chosen to satisfy conditions that ensure suitable convergence \citep{armijo1966minimization,nocedal2006numerical}. Stochastic versions of these line searches were proposed by \citet{mahsereci2017probabilistic} and \citet{vaswani2019painless}. An advantage, in terms of simplicity, of our framework over these methods is that it only requires a single additional function evaluation, keeping the iteration cost comparably low. 

% The parameter update in Eq.~\eqref{eq:mean_update} looks similar to the update of the Kalman filter, which is an instance of Gaussian inference. Our algorithm does not make the assumptions about the temporal evolution of the observations that are central to the Kalman filter, nor do we track the posterior uncertainty to maintain the connection to the optimization algorithms. Further parallels can be drawn to Steepest Descent methods \citep[\textsection 9.4]{boyd2004convex} where optimization is done w.r.t. a general metric, in our case the inverse covariance matrix. 

The Polyak (\citeyear{polyak87optimization}) step, which is a special case of our probabilistic treatment, has previously been used in machine learning for models that satisfy \textit{interpolation} \citep{loizou2020stochasticpolyak}.
% that can reach a training loss close to zero on the assigned problem. This property is often referenced as \textit{interpolation} due to the model being able to interpolate between parameter constellations that yield optimal training loss. 
\citet{loizou2020stochasticpolyak} used the Polyak step together with SGD and proved a convergence rate for the algorithm. Around the same time \citet{berrada2019alig} proposed the ALI-G algorithm which also amounts to a stochastic Polyak step for SGD and a version that incorporates a form of momentum update.

The L4 optimizer of \citet{rolinek2018l4} estimates $f*$ and uses the Polyak step to train deep models. Compared to our algorithm it relies on different estimators for the gradient to specifically speed up Momentum and Adam. It also avoids the function re-evaluation but instead introduces additional hyperparameters to estimate the lower bound $f^*$, making it more sensitive to varying problem setups, cf.~results in Fig.~\ref{fig:deepobs}. 
Such an estimate is also possible to include in our algorithm but was not considered further but instead we focused on the scaling of the covariance.
% during training and properly scale each update. L4 avoids the costly re-evaluation but instead requires additional tuning between problems.
%It is however difficult to find an estimate of $f^*$ for general problems.

Another similar line of work is that of \citet{vaswani2020armijols} who extend the line search of \citet{vaswani2019painless} and the Polyak step of \citet{loizou2020stochasticpolyak} for problems that satisfy interpolation. The main contribution was to use a general metric to recover additional optimization algorithms (the diagonal versions in Tab.~\ref{tab:optimizers}) and analyze the convergence properties. Compared to our work it does not consider the connection to probabilistic inference nor the additional optimizers. It is similar to this work in the sense that Gaussian inference also uses a general metric induced by the inverse covariance matrix. Moreover, we do not specifically consider the interpolation setting but instead aim at adapting the learning rate for general problems. The usage of a line search introduces the need for $\ge1$ additional function evaluations per batch whereas ours rely on a single re-evaluation.

The derivations of Sec.~\ref{sec:method} are reminiscent of probabilistic linear algebra routines with additional noise \citep{hennig2015pn,deroos2019active,cockayne2019bayesian}. Our algorithm could operate in a similar manner if the same batch and Hessian is used for repeated parameter updates and the posterior covariance is propagated. Instead we focused on the connection to first-order optimization algorithms for large-scale machine learning tasks.

% \citet{Schaul2012_no_more_pesky_rate,Zeiler2012_adadelta} both used gradient information to estimate a diagonal approximation of the curvature to decide a suitable step length per parameter. 
% Our approach does not make any assumptions about the Hessian and requires less information, and is therefore more general. 
% Several effective optimization algorithms make use of the same approximate curvature information to decide the next search direction, and our developments can in turn be used on top of this to determine the next step length.

% A related but complementary line of work was presented but \citet{balles2017coupling} who advocate adaptation of the batch size to account for the presence of noise over learning rate decay. Our work is complementary in the sense that it tries to adapt the learning rate to fit the observations and noise.  

\section{CONCLUSION}% AND FUTURE WORK}
\label{sec:conclusion}
We have proposed an algorithm motivated by Gaussian inference, to construct a family of update rules that perform a learning rate adaptation for popular first order stochastic optimization routines. 
The algorithm is applicable to optimization routines where the step direction can be phrased as the product of a symmetric, positive definite matrix with the gradient. 
It uses a local quadratic approximation of the loss function defined by the underlying optimization algorithm to adaptively scale the step size.
In our experiments, the algorithm is able to efficiently adapt the learning rate across several initial learning rates, optimizers and deep learning problems. The robust algorithm achieves competitive performance compared to hand-tuned learning rates, Hypergradient descent and the L4 optimizer with less tuning required. 
The proposed adaptation scheme thus offers a way to automatically update the learning rate of deep learning optimizers within the inner loop -- removing the need for outer-loop parameter tuning of the learning rate which comes at high cost in terms of human labor and hardware resources. 
% The probabilistic interpretation also opens up new research directions regarding the lower bound or uncertainty that could speed up the inference even further.

\newpage

% \begin{contributions} % will be removed in pdf for initial submission,
%                       % so you can already fill it to test with the
%                       % ‘accepted’ class option
%     Briefly list author contributions.
%     This is a nice way of making clear who did what and to give proper credit.

%     H.~Q.~Bovik conceived the idea and wrote the paper.
%     Coauthor One created the code.
%     Coauthor Two created the figures.
% \end{contributions}

% \begin{acknowledgements} % will be removed in pdf for initial submission,
%                          % so you can already fill it to test with the
%                          % ‘accepted’ class option
%     Briefly acknowledge people and organizations here.

%     \emph{All} acknowledgements go in this section.
% \end{acknowledgements}

\bibliography{bibliography}

\clearpage
% \newpage
\appendix

\include{supplement}

\end{document}

%% file: supplement.tex
\begin{figure*}[ht]
\centering
\begin{subfigure}{.33\textwidth}
  \centering
  \includegraphics{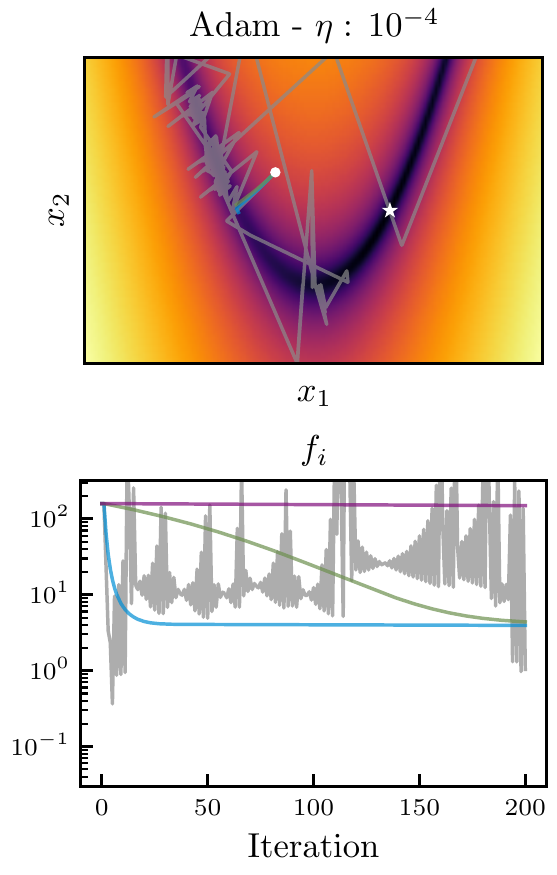}
  \caption{Fixed $\eta$}
  \label{fig:fixed_small}
\end{subfigure}%
\begin{subfigure}{.33\textwidth}
  \centering
  \includegraphics{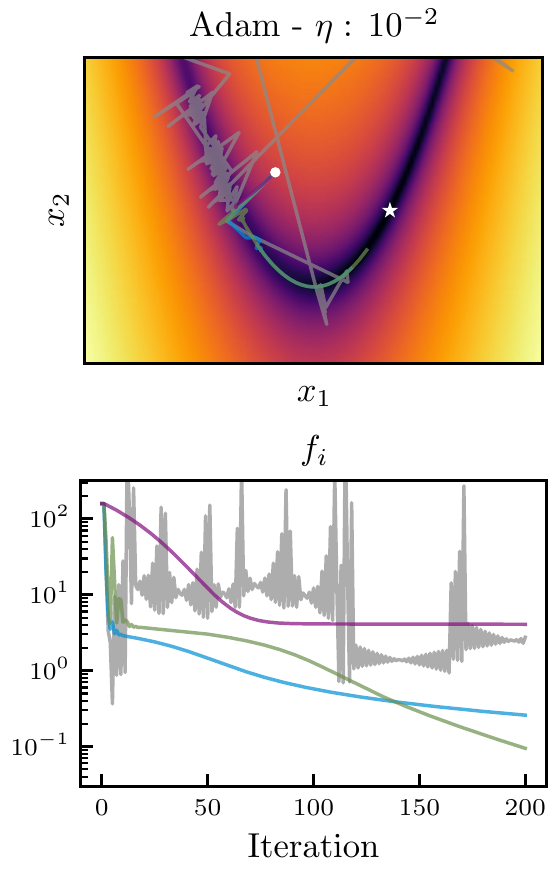}
  \caption{Fixed $\eta$}
  \label{fig:fixed_large}
\end{subfigure}%
\begin{subfigure}{.33\textwidth}
  \centering
  \includegraphics{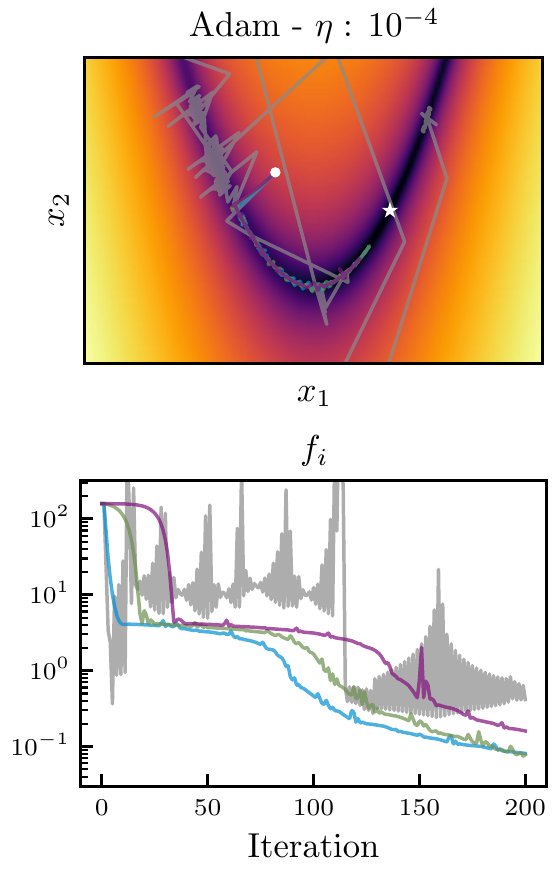}
  \caption{Adapted $\eta$}
  \label{fig:adapt_small}
\end{subfigure}%
\caption{Influence of variance (learning rate) adaption for inference step on 2-d Rosenbrock (\citeyear{rosenbrock1960function}) function. Each figure shows the standard step of the optimizer (\protect\figline{purpleC}),
 the inference with $R_i=0$, i.e., Polyak step (\protect\figline{grayC}), inference with fixed $R_i=0.1$ (\protect\figline{blueC}) and the inference with adaptive $R_i=0.05f_i$ (\protect\figline{greenC}). Each run used the Adam optimizer with $\beta_1=0.7$, $\beta_2=0.999$ and starting learning rate indicated in the figure title. Each of the inference steps use the correct $f^*=0$. The gray line uses $R_i=0$ and is therefore agnostic to the learning rate and should result in the same iterates for all three setups. This is the case up to approximately iteration 100 after which they deviate.
  }
\label{fig:rosenbrock}
\end{figure*}

\section{PROBABILISTIC MODEL}
\label{A:probabilistic_model}
The main parameters of the probabilistic model ($f^*$, $R_i$, $\eta$) have a complicated interplay which affects the \textit{modus operandi} of the algorithm, depending on available information. Fig.~\ref{fig:rosenbrock} highlights some of the difficulties related to setting the parameters $\eta$, $f^*$ and $R_i$ in the probabilistic update. If the global lower bound $f^*$ is known (0 in this case) it is still not straightforward to properly set $\eta$ w.r.t. $R_i$. Adapting the variance on-the-go alleviates this problem when $R_i>0$ and when the standard step of the underlying optimizer is used. One could just as easily consider an algorithm where $f^*$ is estimated and provided externally instead of $R_i$.

\clearpage
\section{ADDITIONAL EXPERIMENTS}
\label{A:additional}
\begin{figure*}[ht]
\centering
\includegraphics{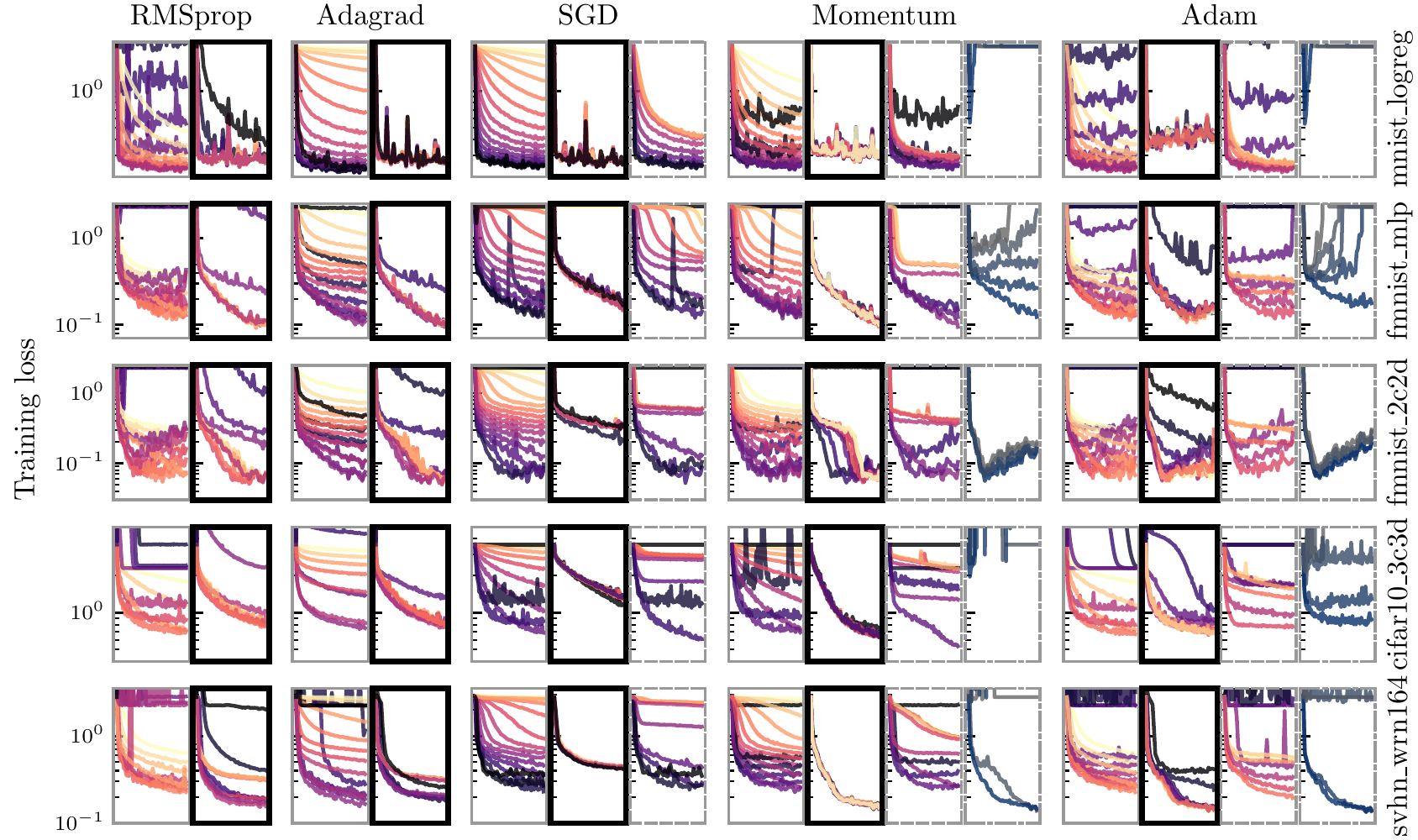}
\caption{\label{fig:deepobs_train_loss}Training loss for the experiments presented in the main part of the paper. For details see Sec.~\ref{A:additional}.}
\end{figure*}

\begin{figure*}[ht]
\centering
\includegraphics{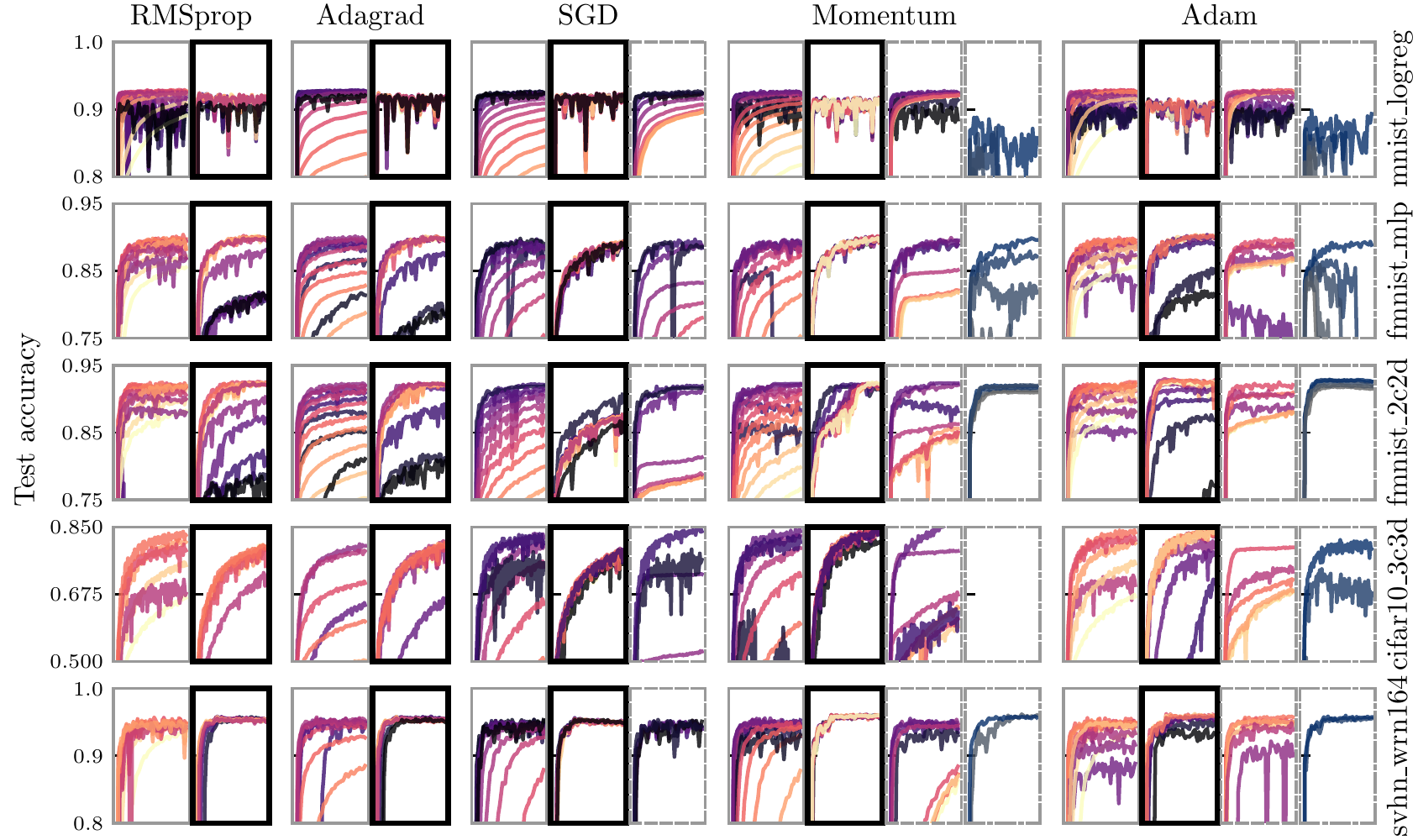}
\caption{\label{fig:deepobs_test_acc}Test accuracy for the experiments presented in the main part of the paper. For details see Sec.~\ref{A:additional}.}
\end{figure*}

The experiments are presented in the same way as the results of the paper. Meaning that each figure shows a loss metric for different test problems (rows) and optimizers (columns). The leftmost graph in each group (thin gray border) has a fixed learning rate and next to it (thick black border) is the proposed adaptation. A dashed border indicates results for hypergradient descent \cite{baydin2017automatic} and the dash-dotted are results for L4 \cite{rolinek2018l4}. 
Each graph contains experiments with initial learning rates indicated by colors in the range $10^{-5}$ (\protect\figline{lowcolor}) through $10^{-3}$ (\protect\figline{midcolor}) to $10^{0}$ (\protect\figline{highcolor}). In the case of L4 the learning is replaced with $\alpha_{L4}$ values between 0.15 (\protect\figline{l4lowcolor}) and 0.25 (\protect\figline{l4highcolor}). In every problem each optimizer ran for 50 epochs except for cifar10\_3c3d which ran for 100 epochs. All hyperparameters of the optimizers were left at the default values except for the momentum term of the proposed adaptation which was set to 0.5 instead of default 0.9 for Momentum and Adam. Here we additionally include the results of RMSprop and Adagrad in the comparison. The training loss of the experiments in the main paper are available in Fig.~\ref{fig:deepobs_train_loss} and a zoomed-in version of the test accuracy can be seen in Fig.~\ref{fig:deepobs_test_acc}.

\newpage
\subsection{CIFAR-100} % (fold)
\begin{figure*}[ht]
\centering
\includegraphics{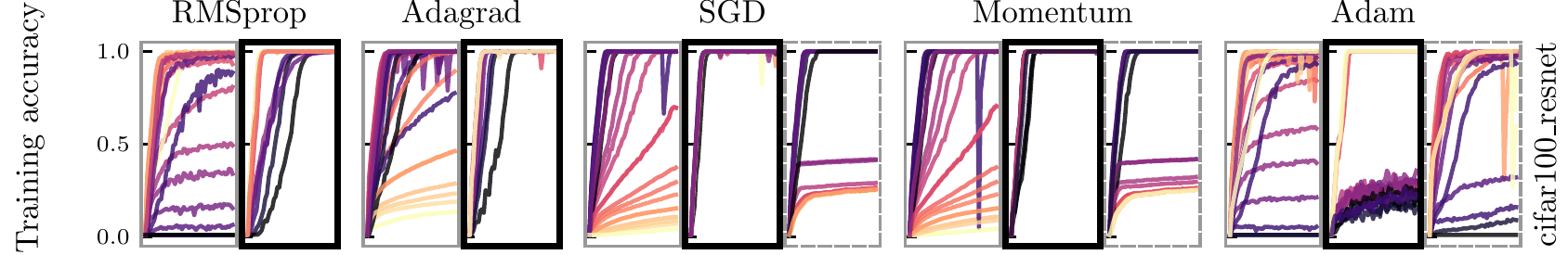}\\
\includegraphics{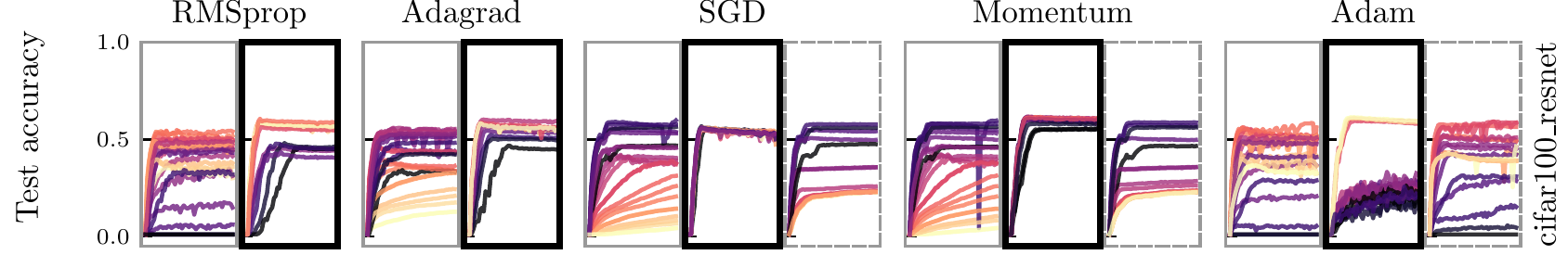}\\
\includegraphics{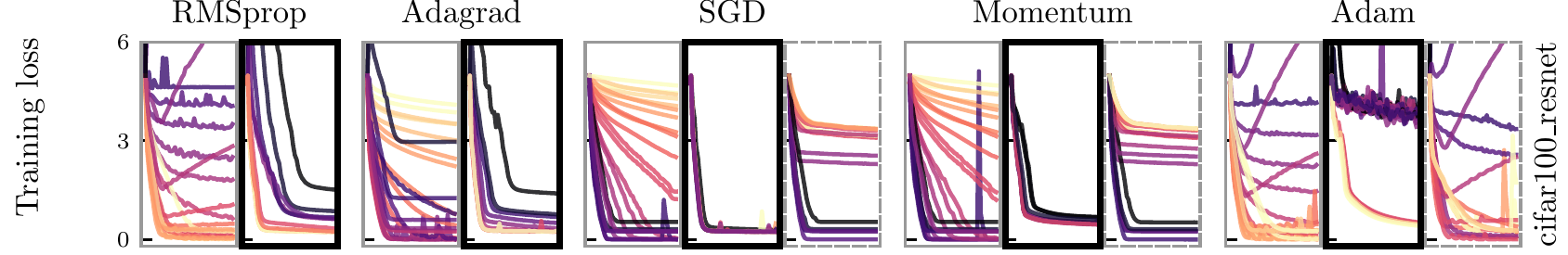}\\
\includegraphics{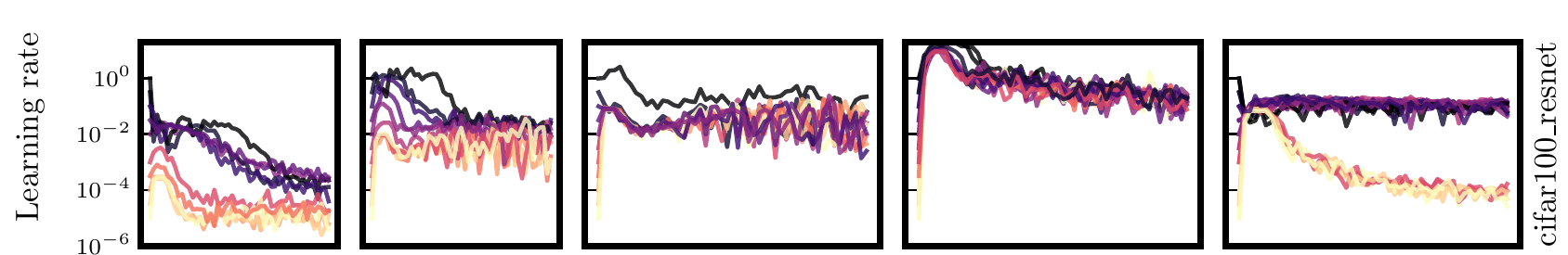}\\
\caption{\label{fig:cifar100} Results for a model trained on CIFAR-100 for 50 epochs. All the settings and limits are the same as the results from DeepOBS. The last row shows the learning rate that was used at the end of each epoch for the proposed learning rate adaptation.}
\end{figure*}

\label{A:cifar100}
To test the optimization on a larger model and dataset we used the ResNet18 implementation from the Pytorch model zoo and trained the model on the CIFAR-100 dataset with a batch size of 128. The used $l_2$-regularization of $5\cdot10^{-4}$ was too low for the model which resulted in overfitting and poor generalization performance, but the overall trend compared to the problems from DeepOBS is still visible. In Fig.~\ref{fig:cifar100} we see different metrics evolve during the training and the learning rate. For each of the optimizers there seems to be an initial convergence point for the learning rate that then transitions into a more noisy regime. 

%% file: article.bbl
\begin{thebibliography}{35}
\providecommand{\natexlab}[1]{#1}
\providecommand{\url}[1]{\texttt{#1}}
\expandafter\ifx\csname urlstyle\endcsname\relax
  \providecommand{\doi}[1]{doi: #1}\else
  \providecommand{\doi}{doi: \begingroup \urlstyle{rm}\Url}\fi

\bibitem[Andrychowicz et~al.(2016)Andrychowicz, Denil, Gomez, Hoffman, Pfau,
  Schaul, Shillingford, and De~Freitas]{andrychowicz2016learn2learn}
Marcin Andrychowicz, Misha Denil, Sergio Gomez, Matthew~W Hoffman, David Pfau,
  Tom Schaul, Brendan Shillingford, and Nando De~Freitas.
\newblock Learning to learn by gradient descent by gradient descent.
\newblock In \emph{Advances in neural information processing systems (NIPS)},
  pages 3981--3989, 2016.

\bibitem[Armijo(1966)]{armijo1966minimization}
Larry Armijo.
\newblock Minimization of functions having lipschitz continuous first partial
  derivatives.
\newblock \emph{Pacific Journal of mathematics}, 16\penalty0 (1):\penalty0
  1--3, 1966.

\bibitem[Asi and Duchi(2019)]{asi2019importance}
Hilal Asi and John~C Duchi.
\newblock The importance of better models in stochastic optimization.
\newblock \emph{Proceedings of the National Academy of Sciences}, 116\penalty0
  (46):\penalty0 22924--22930, 2019.

\bibitem[Baydin et~al.(2017)Baydin, Pearlmutter, Radul, and
  Siskind]{baydin2017automatic}
At{\i}l{\i}m~G{\"u}nes Baydin, Barak~A Pearlmutter, Alexey~Andreyevich Radul,
  and Jeffrey~Mark Siskind.
\newblock Automatic differentiation in machine learning: a survey.
\newblock \emph{The Journal of Machine Learning Research}, 18\penalty0
  (1):\penalty0 5595--5637, 2017.

\bibitem[Baydin et~al.(2018)Baydin, Cornish, Rubio, Schmidt, and
  Wood]{baydin2017hyper}
Atilim~Gunes Baydin, Robert Cornish, David~Martinez Rubio, Mark Schmidt, and
  Frank Wood.
\newblock Online learning rate adaptation with hypergradient descent.
\newblock In \emph{International Conference on Learning Representations
  (ICLR)}, 2018.

\bibitem[Berrada et~al.(2019)Berrada, Zisserman, and Kumar]{berrada2019alig}
Leonard Berrada, Andrew Zisserman, and M~Pawan Kumar.
\newblock Training neural networks for and by interpolation.
\newblock \emph{arXiv preprint arXiv:1906.05661}, 2019.

\bibitem[Cockayne et~al.(2019)Cockayne, Oates, Ipsen, Girolami,
  et~al.]{cockayne2019bayesian}
Jon Cockayne, Chris~J Oates, Ilse~CF Ipsen, Mark Girolami, et~al.
\newblock A {B}ayesian conjugate gradient method.
\newblock \emph{Bayesian Analysis}, 14, 2019.

\bibitem[Dauphin et~al.(2015)Dauphin, de~Vries, Chung, and Bengio]{Dauphin2015}
Yann~N Dauphin, Harm de~Vries, Junyoung Chung, and Yoshua Bengio.
\newblock Rmsprop and equilibrated adaptive learning rates for non-convex
  optimization.
\newblock In \emph{ICML workshop on Deep learning}, 2015.

\bibitem[de~Roos and Hennig(2019)]{deroos2019active}
Filip de~Roos and Philipp Hennig.
\newblock Active probabilistic inference on matrices for pre-conditioning in
  stochastic optimization.
\newblock In \emph{The 22nd International Conference on Artificial Intelligence
  and Statistics}, volume~89 of \emph{Proceedings of Machine Learning
  Research}. PMLR, 2019.

\bibitem[Duchi et~al.(2011)Duchi, Hazan, and Singer]{Duchi2011_adagrad}
John Duchi, Elad Hazan, and Yoram Singer.
\newblock Adaptive subgradient methods for online learning and stochastic
  optimization.
\newblock \emph{Journal of Machine Learning Research}, 2011.

\bibitem[Goodfellow et~al.(2016)Goodfellow, Bengio, and
  Courville]{Goodfellow-et-al-2016}
Ian Goodfellow, Yoshua Bengio, and Aaron Courville.
\newblock \emph{Deep Learning}.
\newblock MIT Press, 2016.

\bibitem[Hardt et~al.(2016)Hardt, Recht, and Singer]{hardt2016train}
Moritz Hardt, Ben Recht, and Yoram Singer.
\newblock Train faster, generalize better: Stability of stochastic gradient
  descent.
\newblock In \emph{International Conference on Machine Learning}. PMLR, 2016.

\bibitem[{He} et~al.(2016){He}, {Zhang}, {Ren}, and {Sun}]{He_resnet}
Kaiming {He}, Xiangyu {Zhang}, Shaoqing {Ren}, and Jian {Sun}.
\newblock Deep residual learning for image recognition.
\newblock In \emph{IEEE Conference on Computer Vision and Pattern Recognition
  (CVPR)}, pages 770--778, 2016.

\bibitem[Hennig et~al.(2015)Hennig, Osborne, and Girolami]{hennig2015pn}
Philipp Hennig, Michael~A Osborne, and Mark Girolami.
\newblock Probabilistic numerics and uncertainty in computations.
\newblock \emph{Proceedings of the Royal Society A: Mathematical, Physical and
  Engineering Sciences}, 471\penalty0 (2179):\penalty0 20150142, 2015.

\bibitem[Kingma and Ba(2014)]{Kingma2014_Adam}
Diederik Kingma and Jimmy Ba.
\newblock Adam: A method for stochastic optimization.
\newblock \emph{International Conference on Learning Representations (ICLR)},
  2014.

\bibitem[Krizhevsky(2009)]{Krizhevsky09learningmultiple}
Alex Krizhevsky.
\newblock Learning multiple layers of features from tiny images.
\newblock Master's thesis, Department of Computer Science, University of
  Toronto, 2009.

\bibitem[Loizou et~al.(2020)Loizou, Vaswani, Laradji, and
  Lacoste-Julien]{loizou2020stochasticpolyak}
Nicolas Loizou, Sharan Vaswani, Issam Laradji, and Simon Lacoste-Julien.
\newblock Stochastic polyak step-size for sgd: An adaptive learning rate for
  fast convergence.
\newblock \emph{arXiv preprint arXiv:2002.10542}, 2020.

\bibitem[Mahsereci and Hennig(2017)]{mahsereci2017probabilistic}
Maren Mahsereci and Philipp Hennig.
\newblock Probabilistic line searches for stochastic optimization.
\newblock \emph{The Journal of Machine Learning Research}, 18\penalty0
  (1):\penalty0 4262--4320, 2017.

\bibitem[Netzer et~al.(2011)Netzer, Wang, Coates, Bissacco, Wu, and
  Ng]{netzer2011svhn}
Yuval Netzer, Tao Wang, Adam Coates, Alessandro Bissacco, Bo~Wu, and Andrew~Y
  Ng.
\newblock Reading digits in natural images with unsupervised feature learning.
\newblock In \emph{NIPS Workshop on Deep Learning and Unsupervised Feature
  Learning}, 2011.

\bibitem[Nocedal and Wright(2006)]{nocedal2006numerical}
Jorge Nocedal and Stephen Wright.
\newblock \emph{Numerical optimization}.
\newblock Springer Science \& Business Media, 2006.

\bibitem[Paszke et~al.(2019)Paszke, Gross, Massa, and et. al.]{paszke19pytorch}
Adam Paszke, Sam Gross, Francisco Massa, and et. al.
\newblock Pytorch: An imperative style, high-performance deep learning library.
\newblock In \emph{Advances in Neural Information Processing Systems
  (NeurIPS)}, pages 8024--8035, 2019.

\bibitem[Polyak(1987)]{polyak87optimization}
Boris Polyak.
\newblock \emph{Introduction to Optimization}.
\newblock 1987.

\bibitem[Riedmiller and Braun(1992)]{riedmiller1992rprop}
Martin Riedmiller and Heinrich Braun.
\newblock Rprop-a fast adaptive learning algorithm.
\newblock In \emph{Proc. of ISCIS VII)}. Citeseer, 1992.

\bibitem[Robbins and Monro(1951)]{robbins1951stochastic}
Herbert Robbins and Sutton Monro.
\newblock A stochastic approximation method.
\newblock \emph{The annals of mathematical statistics}, 22\penalty0
  (3):\penalty0 400--407, 1951.

\bibitem[Rolinek and Martius(2018)]{rolinek2018l4}
Michal Rolinek and Georg Martius.
\newblock L4: Practical loss-based stepsize adaptation for deep learning.
\newblock In \emph{Advances in Neural Information Processing Systems
  (NeurIPS)}, pages 6433--6443, 2018.

\bibitem[Rosenbrock(1960)]{rosenbrock1960function}
H.~H. Rosenbrock.
\newblock An automatic method for finding the greatest or least value of a
  function.
\newblock \emph{The Computer Journal}, 3, 1960.

\bibitem[Ruder(2016)]{Ruder2016_review}
Sebastian Ruder.
\newblock An overview of gradient descent optimization algorithms.
\newblock Technical report, arXiv:1609.04747, 2016.

\bibitem[Schaul et~al.(2013)Schaul, Zhang, and
  Lecun]{Schaul2012_no_more_pesky_rate}
Tom Schaul, Sixn Zhang, and Yann Lecun.
\newblock No more pesky learning rates.
\newblock In \emph{International Conference on Machine Learning (ICML)}, 2013.

\bibitem[Schneider et~al.(2019)Schneider, Balles, and
  Hennig]{SchBalHen19deepobs}
Frank Schneider, Lukas Balles, and Philipp Hennig.
\newblock Deep{OBS}: A deep learning optimizer benchmark suite.
\newblock In \emph{International Conference on Learning Representations
  (ICLR)}, May 2019.

\bibitem[Sutskever et~al.(2013)Sutskever, Martens, Dahl, and
  Hinton]{Sutskever2013}
Ilya Sutskever, James Martens, George Dahl, and Geoffrey Hinton.
\newblock On the importance of initialization and momentum in deep learning.
\newblock \emph{International Conference on Machine Learning (ICML)}, pages
  1139--1147, 2013.

\bibitem[Vaswani et~al.(2019)Vaswani, Mishkin, Laradji, Schmidt, Gidel, and
  Lacoste-Julien]{vaswani2019painless}
Sharan Vaswani, Aaron Mishkin, Issam Laradji, Mark Schmidt, Gauthier Gidel, and
  Simon Lacoste-Julien.
\newblock Painless stochastic gradient: Interpolation, line-search, and
  convergence rates.
\newblock In \emph{Advances in Neural Information Processing Systems}, pages
  3732--3745, 2019.

\bibitem[Vaswani et~al.(2020)Vaswani, Kunstner, Laradji, Meng, Schmidt, and
  Lacoste-Julien]{vaswani2020armijols}
Sharan Vaswani, Frederik Kunstner, Issam Laradji, Si~Yi Meng, Mark Schmidt, and
  Simon Lacoste-Julien.
\newblock Adaptive gradient methods converge faster with over-parameterization
  (and you can do a line-search).
\newblock \emph{arXiv preprint arXiv:2006.06835}, 2020.

\bibitem[Wu et~al.(2020)Wu, Hu, Xiong, Huan, Braverman, and Zhu]{wu2020noisy}
Jingfeng Wu, Wenqing Hu, Haoyi Xiong, Jun Huan, Vladimir Braverman, and
  Zhanxing Zhu.
\newblock On the noisy gradient descent that generalizes as sgd.
\newblock In \emph{International Conference on Machine Learning}. PMLR, 2020.

\bibitem[Zagoruyko and Komodakis(2016)]{Zagoruyko16WRN}
Sergey Zagoruyko and Nikos Komodakis.
\newblock Wide residual networks.
\newblock In Edwin R.~Hancock Richard C.~Wilson and William A.~P. Smith,
  editors, \emph{Proceedings of the British Machine Vision Conference (BMVC)},
  pages 87.1--87.12. BMVA Press, September 2016.
\newblock ISBN 1-901725-59-6.
\newblock \doi{10.5244/C.30.87}.
\newblock URL \url{https://dx.doi.org/10.5244/C.30.87}.

\bibitem[Zeiler(2012)]{Zeiler2012_adadelta}
Matthew~D Zeiler.
\newblock Adadelta: An adaptive learning rate method.
\newblock Technical report, arXiv:1212.5701, 2012.

\end{thebibliography}
